\documentclass{article}

\PassOptionsToPackage{numbers, compress}{natbib}
\usepackage[preprint]{neurips_2026}

\usepackage[utf8]{inputenc}
\usepackage[T1]{fontenc}
\usepackage{hyperref}
\usepackage{url}
\usepackage{booktabs}
\usepackage{multirow}
\usepackage{amsmath}
\usepackage{amsfonts}
\usepackage{amssymb}
\usepackage{amsthm}
\theoremstyle{plain}
\newtheorem{theorem}{Theorem}
\newtheorem{corollary}{Corollary}
\theoremstyle{definition}
\newtheorem{definition}{Definition}
\theoremstyle{remark}

\usepackage{nicefrac}
\usepackage{microtype}
\usepackage{xcolor}
\usepackage{graphicx}
\graphicspath{{figures_tikz/}}

\title{Stable Agentic Control: Tool-Mediated LLM Architecture for Autonomous Cyber Defense}

\author{%
  \textbf{Kerri Prinos} \quad \textbf{Lilianne Brush} \quad \textbf{Cameron Denton} \quad \textbf{Zhanqi Wang} \\
  \textbf{Joshua Knox} \quad \textbf{Snehal Antani} \quad \textbf{Anton Foltz} \quad \textbf{Amy Villase\~{n}or} \\
  Horizon3.ai \\
  San Francisco, CA \\
  \texttt{\{kerri.prinos, lili.brush, cameron.denton, zhanqi,} \\
  \texttt{joshua.knox, antani, anton.foltz, amy.villasenor\}@horizon3.ai}
}

\begin{document}

\maketitle

\begin{abstract}
Agentic systems involved in high-stake decision-making under adversarial pressure need formal guarantees not offered by existing approaches. Motivated by the operational needs of security operations centers (SOCs) that must configure endpoint detection and response (EDR) policies under adversarial pressure, we present a tool-mediated architecture: LLM agents use deterministic tools (Stackelberg best-response, Bayesian observer updates, attack-graph primitives) and select from finite action catalogs enforced at the tool-output interface. A composite Lyapunov function machine-checked in Lean~4 with zero \texttt{sorry} certifies controllability, observability from asymmetric sensor data, and Input-to-State Stability (ISS) robustness under intelligent adversarial disturbance, with two corollaries extending the certificate to any controller or adversary from the catalogs. On 282 real enterprise attack graphs, the claims hold with margin. On paired offensive/defensive telemetry, a tool-mediated Claude Sonnet~4 controller reduces the attacker's expected payoff (game value) by $59\%$ relative to a deterministic greedy baseline, with zero variance across 40 runs at four temperatures. A Claude Haiku~4.5 controller converges to suboptimal game values but stays catalog-bounded over an additional $40$ runs, demonstrating that architectural stability is not dependent on the controller capability. The LLM agent's non-determinism furthers creative exploration of strategies, while the tool-mediated architecture ensures system stability.
\end{abstract}

\section{Introduction}
\label{sec:introduction}
Agentic AI is redefining the cyber threat landscape. CrowdStrike reported an 89\% increase in attacks by AI-enabled adversaries in their 2026 Global Threat Report~\cite{crowdstrike}. AI introduces new attacks, acts as a force multiplier, and accelerates breakout speed, giving defenders less time to detect and respond to threats~\cite{crowdstrike}. In this new era, there is a critical need for reliable and rapid agentic defense systems that can keep pace with adaptive adversaries. Reinforcement learning and self-play, where autonomous agents learn the optimization of defense strategies through adversarial interaction, is a promising approach to build smarter defensive agents~\cite{hammar}. However, existing approaches converge on optimal strategies in $O(10^3)$ training episodes~\cite{hammar} which is impractical when each episode is a real pentest with operational cost and risk. Autonomous attack and defense agents have independently demonstrated great success in real-world environments~\cite{anthropic,sentinelone_defense}. We envision a system where an LLM-based defense agent meets its match: learning optimal defensive strategies on the fly against an adaptive LLM-based attacker agent. 

Within a feedback control loop, an LLM-based defense agent analyzes attack graphs, selects hardening actions, and invokes simulation tools, operating with what Eslami and Yu~\cite{eslami} characterize as endogenous modification of the control architecture. Simultaneously, an adaptive adversary observes the defender's deployments and best-responds with novel attack strategies, acting as an intelligent disturbance on the system. But these agents are non-deterministic. Identical inputs produce different action selections across runs, with accuracy variance up to $15\%$ even at $T=0$~\cite{atil}. This non-determinism makes it difficult to predict agent behavior, and unconstrained agents can exhibit destructive actions in production~\cite{shapira}. Zhu~\cite{zhu} argues that controllability, observability, and stability are operational necessities for agentic cybersecurity and suggests Lyapunov stability as the required formal guarantee. The question is whether we can provide such guarantees despite the agents' non-determinism.

Several research directions converge on this need: guaranteed-safe-AI frameworks~\cite{dalrymple}, runtime enforcement~\cite{wang,kamath,bhardwaj}, per-action theorem proving~\cite{rashie2026}, and safe RL with Lyapunov constraints~\cite{chow,berkenkamp,as}, but none provide machine-checked closed-loop stability guarantees where the disturbance is itself a best-responding agent. We take a different approach: combining game theory, control theory, and tool-mediated LLM control into a stable architecture where closed-loop stability is a property of the loop, not of the agent.

From the blue team's perspective, formal guarantees must answer three operational questions: \emph{Controllability} --- does each policy deployment make the defense strictly better? \emph{Robustness} --- when a novel technique emerges, how much damage can it cause? \emph{Observability} --- is the team's understanding of its posture accurate? We formalize these as closed-loop stability properties: monotone decrease of the game value (Claim~i), input-to-state stability under adversarial expansion (Claim~ii), and observer convergence from asymmetric sensor data (Claim~iii).

We present a tool-mediated architectural pattern for LLM-in-the-loop adversarial control, validated on autonomous cyber defense. Our contributions are as follows:

\begin{itemize}
    \item \textbf{Tool-mediated architecture.} Our architecture combines a Bayesian observer over an asymmetrically-visible plant, Stackelberg best-response dynamics, double oracle expansion and catalog-membership enforcement at the tool-output interface which confines both controller and adversary to finite action sets. The LLM defense agent (controller) composes deterministic tool outputs, but does not compute them; no component is learned or fine-tuned, and convergence occurs within a single analysis cycle, requiring no training episodes.
    \item \textbf{Formal verification.} Theorem~\ref{thm:closed-loop-stability} establishes three closed-loop properties --- controllability, Input-to-State Stability (ISS) robustness under best-responding disturbance, and observability from asymmetric sensor data --- via a composite Lyapunov function $V(k) = S(k) + \lambda\theta(k)$, with proofs machine-checked in Lean~4 with zero \texttt{sorry}. Two corollaries extend the certificate to any controller and any adversary drawn from the catalogs. To our knowledge, this is the first mechanically-verified closed-loop stability certificate for a tool-mediated LLM controller.
    \item \textbf{Empirical validation on real-world security data.} We empirically validate stability of the adversarial closed-loop system on 282 real enterprise attack graphs from production pentests (161 organizations, 25 industries) where the adversary \emph{aids} belief-truth alignment ($4.7\times$ reduction in $|S - \hat{S}|$), and on paired Horizon3.ai NodeZero offensive telemetry and Microsoft Defender XDR defensive telemetry where a tool-mediated Claude Sonnet~4 controller achieves $S(k_{\text{final}}) = 0.34$ versus $0.84$ for a deterministic greedy baseline at $\sigma = 0.000$ across $40$ runs at four temperatures, while a less capable Claude Haiku~4.5 controller stays catalog-bounded with non-zero $\sigma$, separating architectural stability from achieved game value.
\end{itemize}

\section{Related Work}
\label{sec:related-work}
Our architecture combines four lines of prior work: composite Lyapunov stability, game-theoretic adversarial modeling, Endpoint Detection and Response (EDR) policy optimization, and tool-mediated LLM scaffolding.

\textbf{Composite Lyapunov stability and ISS.} Hayakawa et al.~\cite{hayakawa} proved that a composite Lyapunov function decomposing into plant and estimator terms guarantees partial asymptotic stability. Jiang and Wang~\cite{jiang} established the ISS-Lyapunov equivalence. Zhu and Basar~\cite{zhu_basar} pioneered the integration of game theory with control-theoretic methods for cyber-physical resilience, although their games-in-games principle addresses physical plant dynamics rather than EDR policy optimization. Eslami and Yu~\cite{eslami} proposed a control-theoretic framework for LLM-based agentic system where runtime-adaptive LLM agents are modeled as endogenous modification of the control architecture, but they explicitly identified verification of stability assumptions for complex decision processes involving LLMs as an open problem.

\textbf{Game-theoretic adversarial modeling.} In a Stackelberg security game~\cite{leitmann,sinha}, the defender commits first and the adversary best-responds. Stackelberg security games with partial observability have been addressed by Durkota et al.~\cite{durkota} with Bayesian formulations and Miehling et al.~\cite{miehling} with POMDP-based dynamic defense. Double-oracle methods~\cite{mcmahan,jain} iteratively expand both strategy sets to converge to game equilibria without full enumeration. Zhang et al.~\cite{zhang_double_oracle_AD} demonstrated scalability of the double oracle method to large AD graphs. Network interdiction~\cite{wood} formalizes the game value as the best surviving path probability which we adopt for $S(k)$. Romano and Pavel \cite{romano} use control theory to prove Nash convergence under exogenous disturbances. However, these methods guarantee convergence of the game equilibria, not stability of a closed-loop adversarial system where beliefs are derived from noisy, real-world sensor data.

\textbf{EDR policy optimization.} EDR is a primary defense technology used by enterprise security teams to monitor, detect, and respond to threats on end-user devices \cite{aarness}. Blue teams operating under partial or asymmetric observability must reason about an attacker's behavior based on defensive telemetry and deploy defensive policies.  Enabling every available policy in block mode is infeasible: each carries operational overhead (false positives, prerequisite dependencies, alert triage cost) that the SOC must absorb within a maintenance window. Analysis of 37 EDR vendors in MITRE ATT\&CK evaluations reveals significant coverage variation requiring attack-graph-level correlation~\cite{shen}. Outkin et al.~\cite{outkin} applied game-theoretic resource allocation to MITRE data; subsequent work explores centrality-based~\cite{aleiadeh} and cost-benefit~\cite{zhang_cost_benefit} approaches. These optimize over abstract allocation, not over a real vendor catalog with detect/block tradeoffs, budget constraints, and measured deployment outcomes.

\textbf{Tool-mediated LLM scaffolding.} ReAct~\cite{yao2022react} interleaves reasoning with tool calls; SayCan~\cite{ahn2022saycan} grounds actions through learned value functions; Inner Monologue~\cite{huang2022inner} closes the loop via language feedback. G-CTR~\cite{mayoralvilches} pairs a game-theoretic solver to guide offensive and defensive LLM agents ($5.2\times$ variance reduction); MaMa~\cite{nother2026mama} uses a Stackelberg meta-game to harden multi-agent designs. These reduce variance or secure outputs but do not certify system-level closed-loop stability. 

\section{Approach}
\label{sec:approach}
We cast autonomous cyber defense as a closed-loop control problem blending LLM tool use, game theory, and control-theoretic stability. The system is a discrete-time non-linear feedback loop:

\begin{equation}
  \begin{aligned}
    \mathcal{G}(k+1) &= f\bigl(\mathcal{G}(k), u(k), w(k)\bigr), \\
    y(k) &= h\bigl(\mathcal{G}(k), w(k)\bigr).
  \end{aligned}
  \end{equation}

where $\mathcal{G}(k)$ is the state of the adversarial graph, $u(k)$ the control input (defender actions), $w(k)$ the disturbance (adversary actions), and $y(k)$ the observation (defender telemetry). A Stackelberg double-oracle game under asymmetric observability is integrated into the loop.

\textbf{Plant.} The plant is a directed adversarial graph $\mathcal{G}(k) = (\mathcal{V},\mathcal{E})$: nodes are hosts at a stage of the attack chain (foothold, lateral, objective); edges are attacker actions (MITRE ATT\&CK techniques applied to a host). Edge-local quantities depend on $(e,k)$:
\begin{itemize}
    \item $\mathrm{payoff}(e,k) \in [0,1]$: attacker's stage payoff if edge $e$ is traversed at round~$k$ (technique impact and host criticality).
    \item $\mathrm{block}(e,k) \in [0,1]$: probability the defender blocks traversal of edge~$e$ at time~$k$.
    \item $\mathrm{detect}(e,k) \in [0,1]$: probability the defender observes traversal of edge~$e$ at time~$k$.
    \item $P_e(k) \in [0,1]$: defender's posterior uncertainty on edge~$e$ at time~$k$ (belief error between ground truth $\mathcal{G}(k)$ and belief $\hat{\mathcal{G}}(k)$ on that edge).
\end{itemize}
Edges derive from temporal ordering within each host, cross-host credential flow, and causal parent-child links in the pentest attack-chain data. Nodes and edges update at each step via defender and attacker actions.

\textbf{Asymmetric visibility.} The attacker has full visibility of $\mathcal{G}(k)$; the defender maintains a belief graph $\hat{\mathcal{G}}(k)$ built from defensive telemetry. Edges matched to alerts are initialized with uncertainty $P_e = 0.15$; unmatched edges are absent from $\hat{\mathcal{G}}$ --- these ``dark edges'' remain unknown to the defender until ground truth is revealed via the plant transition. The defender anticipates the attacker's best-response over $\hat{\mathcal{G}}(k)$ and may propose blocking, detection, or logging actions to shrink the dark-edge set.

\textbf{Controller.} In our control feedback loop, the defender acts as the controller. Following the Stackelberg game model, the defender is the leader and the attacker observes the defender's strategy and best-responds. Strategy-wise, the defender upgrades existing policies from logging to enforce, or uses the oracle to expand to new policies from a finite catalog $\mathcal{C} = \{p_1, \ldots, p_n\}$ spanning endpoint, identity, and cloud domains (compiled from public vendor and MITRE sources; Appendix~\ref{app:benchmark-datasets}). Each deployment consumes one slot from a per-round budget $B$. The defender selects blocking actions to minimize the attacker's payoff on its belief state graph $\hat{\mathcal{G}}(k)$. The attacker's expected payoff or game value is given by the maximum payoff reachable via any surviving path from ENTRY to OBJECTIVE~\cite{wood,zhang_double_oracle_AD}:

\begin{equation}
  \label{eq:game-value}
    S(k) = \max_{p \in \mathcal{P}}
      \left[
        \biggl( \prod_{e \in p} \bigl(1 - \mathrm{block}(e,k)\bigr) \biggr)
        \cdot \max_{e' \in p} \mathrm{payoff}(e',k)
      \right]
  \end{equation}

where $\mathcal{P}$ is the set of ENTRY-to-OBJECTIVE paths at round~$k$; the product is the path's survival probability and the inner $\max$ is the largest stage payoff on~$p$.

\textbf{Tool-mediated action selection.} The LLM controller never reads $\mathcal{G}(k)$ or $\hat{\mathcal{G}}(k)$ directly, nor computes $S(k)$, enumerates paths, or runs observer updates. All such quantities are produced by a bounded set of deterministic tools implementing Stackelberg best-response, Bayesian observer updates, and attack-graph primitives. The LLM issues tool calls, consumes structured outputs, and selects one catalog action per deployment slot. The defender is exposed to $9$ tools (e.g.\ \texttt{compute\_v\_after\_deploy}, \texttt{simulate\_round\_ahead}, \texttt{get\_critical\_path}, \texttt{identify\_dark\_edges}, \texttt{list\_deployable\_policies}) and the adversary to a mirror $11$-tool suite (e.g.\ \texttt{find\_weakest\_path}, \texttt{evaluate\_new\_edge}, \texttt{find\_zero\_day\_opportunity}); full inventories in Appendix~\ref{app:llm-config}. This follows G-CTR's~\cite{mayoralvilches} pattern (a ReAct~\cite{yao2022react} loop guided by a game-theoretic digest): non-determinism is confined to tool-composition and action-selection over deterministic outputs. Assumption~A2 (catalog finiteness) is enforced at the tool-output interface, so off-catalog proposals are rejected as no-ops. \S\ref{sec:exp2} also evaluates a deterministic greedy controller (no tool calls, catalog policies ranked by marginal $S(k)$ reduction) to isolate what the LLM's tool-composition loop adds over the best deterministic alternative.

\textbf{Disturbance.} The attacker observes the defender's actions and ground truth graph state $\mathcal{G}(k)$ and uses an oracle to best-respond to $\mathcal{G}(k+1) = f(\mathcal{G}(k),u(k))$ by proposing a new edge $e_{\text{new}}$ from a finite attack technique catalog $\mathcal{T}$ to maximize its payoff. New edges connect existing intermediate nodes (no ENTRY $\to$ OBJECTIVE bypass).

\textbf{Observer.} A scalar Kalman filter per edge $e \in \hat{\mathcal{G}}(k)$ contracts uncertainty $P_e$ toward ground truth:
\begin{equation}
\label{eq:kalman-edge-update}
  \begin{aligned}
    K_e &= \frac{P_e(k-1)}{P_e(k-1) + R_k}, \\
    \hat{P}_e(k) &\leftarrow \hat{P}_e(k-1) + K_e\bigl(z_e - \hat{P}_e(k-1)\bigr), \\
    P_e(k) &\leftarrow (1-K_e)\,P_e(k-1).
  \end{aligned}
\end{equation}
Here, $z_e$ is the measurement from matched telemetry or ground-truth reveal, $R_k > 0$ the measurement-noise variance, and $K_e \in (0,1)$ the Kalman gain. Let $E_{\text{obs}}(k) \subseteq E(\hat{\mathcal{G}}(k))$ denote edges that receive a measurement at round~$k$; contraction applies to each $e \in E_{\text{obs}}(k)$ (Theorem~\ref{thm:closed-loop-stability}).

The per-edge innovation measures the gap between prediction and ground truth on an observed edge:
\begin{equation}
\label{eq:innov-edge}
  \text{innov}(e,k) = \bigl(1 - P_e(k)\bigr) \cdot
  \frac{|\Delta\text{detect}(e,k)| + |\Delta\text{block}(e,k)| + |\Delta\text{traversal}(e,k)|}{3}.
\end{equation}
Each $\Delta$-term is the absolute mismatch between belief and revealed ground truth on edge~$e$ (traversal coded $1$/$0$). The factor $(1-P_e(k))$ increases the weight of high-confidence mismatches.

\textbf{Convergence criterion.} We terminate the loop when either of the following conditions is met:
\begin{itemize}
  \item Strong Stackelberg Equilibrium (SSE) \cite{leitmann} criterion is met --- neither the defender nor the attacker can improve their payoff by unilaterally deviating from their current strategy.
  \item The \emph{mean innovation} $\overline{\mathrm{innov}}(k) = \frac{1}{|E(\hat{\mathcal{G}}(k))|}\sum_{e \in E(\hat{\mathcal{G}}(k))} \text{innov}(e,k) < \varepsilon_{\text{innov}}$ for two consecutive rounds, where $\varepsilon_{\text{innov}} > 0$ is a fixed convergence threshold (value in \S\ref{sec:datasets}), indicating belief has converged to ground truth.
\end{itemize}

\S\ref{sec:closed-loop} formalizes the three closed-loop properties --- Controllability, Robustness (ISS), and Observability--- with two corollaries extending them to arbitrary controllers and adversaries. \S\ref{sec:experiments} empirically validates each claim.

\section{Formal Verification of Closed-Loop Stability}
\label{sec:closed-loop}

Proof sketches appear in Appendix~\ref{app:formal}; the full Lean~4 source (five files, ${\sim}300$ lines, zero \texttt{sorry}) is included in the supplementary material.

\textbf{Assumptions.}
\begin{itemize}
  \item[(A1)] The graph $G$ is finite.
  \item[(A2)] The defender's policy catalog $\mathcal{C}$ is finite.
  \item[(A3)] The attacker's new edges are drawn from a finite attack technique set $\mathcal{T}$ and connect existing nodes.
  \item[(A4)] \emph{Persistent deployment}: The defender and the attacker maintain the existing graph structure, update existing edges, or add new edges to the graph. They do not roll back or undo their actions.
  \item[(A5)] The Bayesian observer update is contractive: each observation strictly reduces $P_e$.
\end{itemize}
\smallskip
\begin{definition}[Composite Lyapunov function]
\label{def:composite-lyapunov}
We define a composite Lyapunov function $V(k)$ as the sum of the game value $S(k)$ and a weighted aggregate of edge uncertainties $P_e(k)$, decomposing into plant and estimator terms as in \cite{hayakawa}:
\begin{equation}
\label{eq:lyapunov-function}
  V(k) = S(k) + \lambda \theta(k), \quad \lambda > 0,
\end{equation}

where $S(k)$ is the game value from~\eqref{eq:game-value} and $\theta(k) = \bar{P}_e(k) = \frac{1}{|E(\hat{\mathcal{G}}(k))|}\sum_{e \in E(\hat{\mathcal{G}}(k))} P_e(k) \in [0,1]$ is the mean per-edge posterior uncertainty on the belief graph.
Both $S(k)$ and $\theta(k)$ are positive definite decrescent functions bounded in $[0,1]$. The Lyapunov function $V(k)$ is non-negative and equals zero if and only if no attacker path from ENTRY to OBJECTIVE survives \emph{and} the observer is perfectly calibrated ($P_e = 0$ for every edge in the belief graph). The three claims below are machine-checked in Lean~4 (Appendix~\ref{app:formal}).
\end{definition}
\smallskip
\begin{theorem}[Closed-loop stability]
\label{thm:closed-loop-stability}
Under Assumptions~(A1)--(A5), the closed-loop system satisfies the following:
\begin{enumerate}
  \item[(i)] \textbf{Controllability (monotone decrease toward optimal defense).} When no adversary disturbance occurs,
  \begin{equation}
  \label{eq:composability-monotone}
  V(k+1) \leq V(k) - \alpha_B\bigl(S(k)\bigr) - \lambda\delta\bigl|E_{\text{obs}}(k)\bigr|
  \end{equation}
  where $\alpha_B$ is a class-$\mathcal{K}$ function representing the minimum $S(k)$ reduction from $B$ greedy policy deployments and $\delta > 0$ is the minimum reduction in $\theta$ per observed edge from the Bayesian observer update.
  \item[(ii)] \textbf{Robustness (ISS under adversarial disturbance).} Under adversary graph expansion,
  \begin{equation}
  \label{eq:robustness-iss}
    V(k+1) \leq V(k) - \alpha_B\bigl(S(k)\bigr) + \gamma\bigl(|w(k)|\bigr) - \lambda\delta\bigl|E_{\text{obs}}(k)\bigr|
  \end{equation}
  where $E_{\text{new}}(k)$ denotes the set of edges the attacker adds at round~$k$ (so $\lvert E_{\text{new}}(k)\rvert = n_{\text{new}}$), with $\gamma(|w(k)|) = 0$ if $E_{\text{new}}(k) = \emptyset$ and otherwise
  \begin{equation*}
    \gamma\bigl(|w(k)|\bigr) = (1 - \varepsilon_{\text{antic}})\!\max_{e \in E_{\text{new}}(k)} \mathrm{payoff}(e,k),
  \end{equation*}
  which bounds the single-round $S$ increase from those edges; $\varepsilon_{\text{antic}} \in [0,1]$ is the anticipatory defense effectiveness. The system is ISS when $\alpha_B + \lambda\delta\bigl|E_{\text{obs}}(k)\bigr| > \gamma(|w(k)|) + \lambda\, n_{\text{new}}\, P_{\max}$, where $P_{\max}$ is the maximum initial uncertainty assigned to a new edge. The non-expansivity bound follows from Hayakawa et al.~\cite{hayakawa}.
  \item[(iii)] \textbf{Observability (observer convergence from asymmetric sensor data).}
  \begin{equation}
  \label{eq:observer-calibration}
    \theta(k) \leq (1-\delta)^k \theta(0) + C_{\text{new}}(k)
  \end{equation}
  with $\theta(k)$ as in~\eqref{eq:lyapunov-function}. The factor $(1-\delta)^k \theta(0)$ decays geometrically in the round index~$k$. $C_{\text{new}}(k)$ is bounded by double oracle termination: the adversary's technique set is finite (A4), the node set is finite (A1), anticipatory defense shrinks the effective pool each round, and diminishing returns ensure termination. After termination, $C_{\text{new}}$ stabilizes and the geometric decay dominates, yielding convergence to a bounded neighborhood.
\end{enumerate}
\end{theorem}
\smallskip
\begin{corollary}[Controller-agnostic guarantee]
\label{cor:controller-agnostic}
Theorem~\ref{thm:closed-loop-stability} applies to any controller selecting from $\mathcal{C}$; the bounds in~\eqref{eq:composability-monotone}--\eqref{eq:observer-calibration} depend only on the action space, not on the controller's decision process. An LLM controller operates within the action space defined by $\mathcal{C}$ and inherits all three stability guarantees.
\end{corollary}
\smallskip
\begin{corollary}[Adversary-agnostic bound]
\label{cor:adversary-agnostic}
The disturbance bound $\gamma$ scales with the maximum stage payoff among the edges in $E_{\text{new}}(k)$ and with $(1-\varepsilon_{\text{antic}})$, not with how the adversary ranks the catalog internally. An LLM adversary may discover higher-payoff edges faster than a deterministic one but cannot exceed the maximum payoff over the finite attack technique set (A3)--(A4); the ISS bound holds regardless of the adversary's decision process.
\end{corollary}

\section{Experiments}
\label{sec:experiments}
We validate the architectural pattern along two axes corresponding to the formal results of \S\ref{sec:closed-loop}. \textbf{Experiment~1} tests Claims~(i)--(iii) --- Controllability, Robustness (ISS), and Observability --- on 282 real enterprise attack graphs spanning 161 organizations and 25 industries, using a deterministic greedy controller so that the evidence isolates the contribution of the architecture's deterministic components (finite-catalog action space, Bayesian observer, game-theoretic solver). \textbf{Experiment~2} tests Corollaries~\ref{cor:controller-agnostic} and~\ref{cor:adversary-agnostic} --- controller- and adversary-agnosticism --- on a paired offensive/defensive telemetry, comparing a deterministic greedy controller against two tool-mediated controllers (Claude Sonnet~4 and Claude Haiku~4.5), each swept over four temperatures ($80$ runs total). The cross-LLM contrast tests within-family scaling of the corollaries. None of the controllers have trained or fine-tuned components.

\subsection{Datasets and shared setup}
\label{sec:datasets}
\textbf{Experiment~1.} We construct 282 adversarial graphs from real enterprise penetration tests conducted by Horizon3.ai's NodeZero, a commercial autonomous pentesting platform \cite{nodezero}. From 7{,}372 operations in the data warehouse, 300 are selected via stratified sampling across three complexity bins (Appendix~\ref{app:benchmark-datasets}); 18 are excluded during validation (14 where $S < 0.01$ before any deployment and 4 where no block-mode policy covers any edge). The remaining 282 graphs span 161 customer organizations across 25 industries (healthcare, manufacturing, finance, government). Graphs range from 276 to 3{,}599 edges (median 806). All graphs are sanitized to remove customer-identifiable information (Appendix~\ref{app:benchmark-datasets}). Synthetic sensor alerts at 60\% coverage create realistic asymmetric observation --- approximately 40\% of edges are absent from the defender's belief graph.

\textbf{Experiment~2.} One adversarial graph (23 nodes, 44 edges, 30-action catalog spanning endpoint, identity, and cloud; 66-technique adversary catalog) derived from a pentest of the Game of Active Directory (GOAD)\footnote{\url{https://github.com/Orange-Cyberdefense/GOAD}} environment with paired defensive telemetry (Appendix~\ref{app:goad-environment}). Both defender and adversary observe full ground truth.

\textbf{Shared hyperparameters.} $B = 3$, $R = 0.05$, $\varepsilon_{\text{innov}} = 0.05$, $\varepsilon_V = 10^{-4}$ (Lyapunov convergence threshold, $|V(k){-}V(k{-}1)| < \varepsilon_V$), max 10 rounds, $\lambda = 1.0$, seed~42. Benchmark runs on a single CPU (${\sim}30$~min total); temperature sweep costs \$93.90 total across 80 runs (40 Sonnet-4 + 40 Haiku-4.5). Full justifications in Appendix~\ref{app:hyperparameters}.

\subsection{Experiment 1: Validating Claims (i)--(iii) on 282 real enterprise graphs}
\label{sec:exp1}
\textbf{Setup.} Two conditions per graph: \emph{defender-only} (no adversary disturbance; isolates Claims~(i) Controllability and~(iii) Observability) and \emph{defender~+~attacker} (adaptive adversary injects one new edge per round; exercises all three claims, including~(ii) Robustness/ISS). The controller is deterministic greedy, ranking actions by marginal $S(k)$ reduction. $564$ runs total ($282 \times 2$). Of the 282 graphs, $143$ have at least one catalog-covered high-payoff edge and drive non-trivial $S$ reduction; the remaining $139$ produce flat (still non-increasing) trajectories --- we report effective-$N$ alongside the full sample throughout.

\begin{figure}[!htbp]
  \centering
  \includegraphics[width=\linewidth]{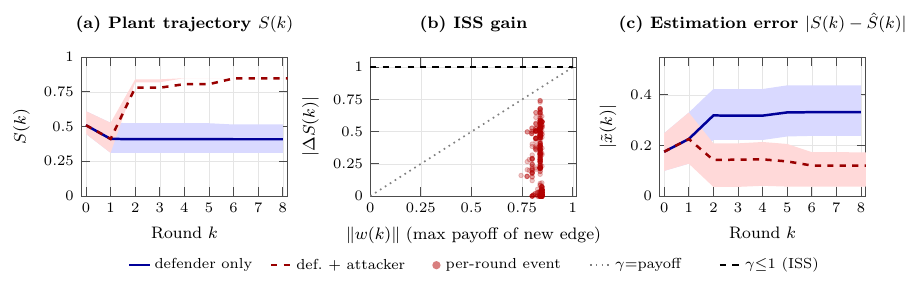}
  \caption{Experiment~1 results on 282 graphs. (a) Plant trajectory $S(k)$: defender-only (blue) monotone $0.51 \to 0.41$; defender+attacker (red) stabilizes at $\approx 0.85$. (b) ISS gain: all 602 disturbance events satisfy $|\Delta S(k)| \leq \gamma = 1.0$; max excursion stays below $0.60$ across all graphs. (c) Belief-truth game-value gap $|S(k)-\hat{S}(k)|$: defender-only plateaus at $0.33$; defender+attacker decreases to $0.12$ as adversary-triggered reveals supply additional Bayesian updates. Final-gap median $0.07$ adversarial vs $0.33$ defender-only ($4.7\times$ improvement).}
  \label{fig:exp1}
\end{figure}

\textbf{Results.} Figure~\ref{fig:exp1} visualizes empirical validation of Claims~(i)--(iii) across all 282 graphs; aggregate metrics per claim are reported inline below.

\textbf{Claim (i): Controllability.} Under defender-only, ground-truth $S(k)$ is non-increasing on every defender turn in $282/282$ graphs (Wilson $95\%$ CI $[0.987, 1.000]$), confirming~\eqref{eq:composability-monotone}. Mean $S$ falls from $0.509$ to $0.408$ ($19.7\%$; bootstrap $95\%$ CI $[17.2\%, 22.3\%]$; paired Wilcoxon $p < 10^{-24}$) in $3.0 \pm 0.1$ rounds, with $97.7\%$ of the reduction in round~1 on the $143/282$ dynamic graphs. Convergence is size-invariant across 276--3{,}599 edges.

\textbf{Claim (ii): Robustness.} Under adversarial expansion ($67.6\%$ of turns), all single-round $S$ spikes satisfy~\eqref{eq:robustness-iss}: max observed $\Delta S = 0.74$, below $\gamma = 1.0$; mean max spike $0.42$ (bootstrap $95\%$ CI $[0.40, 0.44]$; SD $0.14$; Fig.~\ref{fig:exp1}b). Per-graph max excursion stays below $0.60$, a $40\%$ margin to the ISS ceiling. Anticipatory defense under A4 blocks $67/890$ adversary actions outright ($7.5\%$; Wilson $95\%$ CI $[6.0\%, 9.4\%]$).

\textbf{Claim (iii): Observability.} The belief-truth game-value gap $|S(k)-\hat{S}(k)|$ decays geometrically, fitted by $0.064 \cdot 0.10^{k} + 0.007$, a $90\%$ per-round contraction reaching a $0.007$ noise floor by round~2. All 282 graphs converge within 6 rounds with size-invariant decay rate. Counterintuitively, the final gap is $4.7\times$ lower under adversarial pressure (median $0.07$ vs $0.33$; paired Wilcoxon $p < 10^{-32}$, Hodges-Lehmann $\hat{\Delta} = 0.24$, bootstrap $95\%$ CI $[0.22, 0.28]$): adversary-triggered reveals supply additional Bayesian updates, so purple teaming accelerates rather than degrades observability.

\textbf{Composite Lyapunov.} Both components of $V(k) = S(k) + \lambda\theta(k)$ are non-increasing under defender control and bounded under adversarial disturbance; $V(k)$ decreases toward a neighborhood of $V^\ast = 0$ whenever~\eqref{eq:robustness-iss}'s ISS condition holds on all 564 scenarios.

\subsection{Experiment 2: Validating Corollaries 1 and 2 on paired telemetry}
\label{sec:exp2}
\textbf{Setup.} Three controllers on the same GOAD graph with paired Horizon3.ai NodeZero pentest telemetry + Microsoft Defender XDR telemetry (5 hosts, 55-min pentest): \emph{greedy} (deterministic, marginal-$S$ ranking), \emph{Sonnet~4} (tool-mediated with 9 defender tools), and \emph{Haiku~4.5} (identical tool-mediated controller with weaker backbone --- same SDK, prompt, and catalog). The adversary is an LLM with a mirror 11-tool suite (Appendix~\ref{app:llm-config}). Each tool-mediated controller runs $40$ times at four temperatures $\{0.0, 0.3, 0.7, 1.0\}$. Greedy vs Sonnet isolates the LLM's value over a deterministic baseline; the temperature sweep tests Cor.~\ref{cor:controller-agnostic} and Cor.~\ref{cor:adversary-agnostic}; the Sonnet vs Haiku contrast tests within-family scaling.

\textbf{Greedy vs tool-mediated.} Deterministic greedy reaches $S(k_{\text{final}}) = 0.8367$ by deploying 5 catalog policies and plateauing after the adversary's round-1 spike. Sonnet~4 reaches $S(k_{\text{final}}) = 0.3427$ ($59\%$ lower) by deploying the same 5 plus \texttt{mfa\_enforcement} and \texttt{cloud\_app\_security}, breaking the adversary's identity and cloud pivots --- a globally-optimal choice surfaced by \texttt{simulate\_round\_ahead} and \texttt{get\_critical\_path} that greedy's immediate-marginal-value ranking misses. The belief-truth game-value gap shrinks correspondingly: greedy ends with $|S - \hat{S}| = 0.494$ (uncovered adversary expansions persist in the truth graph but are absent from the belief graph) versus $0.0$ for Sonnet~4.

\begin{figure}[!htbp]
  \centering
  \includegraphics[width=\linewidth]{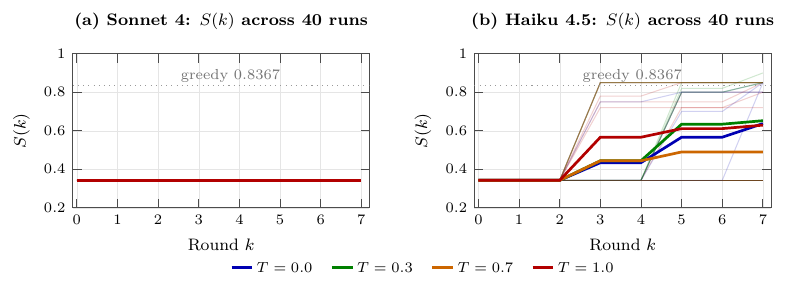}
  \caption{Within-family scaling of LLM stability. (a)~Sonnet~4: all 40 runs converge to $S(k_{\text{final}}) = 0.3427$ with $\sigma = 0$ across temperatures; variance is zero. (b)~Haiku~4.5: same architecture, weaker backbone --- $19/40$ runs reach $S = 0.3427$, $21/40$ stall at $0.85$--$0.90$ ($\sigma = 0.249$, mean $0.603$). Both controllers stay catalog-bounded ($420/420$ deployments on-catalog); the achieved $S$ floor depends on capability, not on the architectural guarantee.}
  \label{fig:llm-stability}
\end{figure}

\textbf{Sonnet~4 sweep (Cor.~\ref{cor:controller-agnostic}, Cor.~\ref{cor:adversary-agnostic}).} Figure~\ref{fig:llm-stability}a shows all $40/40$ Sonnet~4 runs converging to exactly $0.3427$ (Wilson $95\%$ CI $[91.2\%, 100\%]$) --- a literal zero-variance point mass, qualitatively stronger than the variance-reduction typical of LLM-determinism studies (Atil et al.~\cite{atil} report ${\sim}15\%$ accuracy variance at $T{=}0$). Temperature governs \emph{which} action is selected but not the system-level outcome: per-run observer decay rates across the four temperatures show no detectable temperature effect (Kruskal-Wallis $p = 0.17$). Defender Jaccard ranges $0.86$--$0.93$ across temperatures; zero off-catalog hallucinations across $210$ deployments.

\textbf{Haiku~4.5 sweep (within-family scaling).} The same architecture with a less capable backbone exhibits a different pattern (Fig.~\ref{fig:llm-stability}b): $S(k_{\text{final}})$ varies across runs ($\sigma = 0.249$, mean $0.603$, bootstrap $95\%$ CI $[0.527, 0.681]$), with $19/40$ runs reaching $S = 0.3427$ (matching Sonnet~4) and $21/40$ stalling at $0.85$--$0.90$ (greedy-baseline level). Defender Jaccard $0.80$--$0.96$ (comparable to Sonnet~4); zero off-catalog hallucinations across $210$ deployments. Inspection of failure cases shows Haiku~4.5 correctly solves the round-$0$ graph but fails to integrate adversary-expanded edges into its belief graph in subsequent rounds: the inferred $\hat{S}(k)$ stays pinned at the round-$0$ optimum while the ground-truth $S(k)$ climbs as the adversary introduces new techniques ($21/40$ runs exhibit a final belief-truth gap $> 0.1$, Wilson $95\%$ CI $[0.375, 0.671]$). This is a controller-reasoning failure (Haiku does not re-query the belief graph after adversary expansion), not an architectural fault. Mann-Whitney on per-run $S(k_{\text{final}})$ confirms the two controllers differ at $p = 1.6 \times 10^{-7}$.

\textbf{Corollary 1: Controller-agnostic.} Both controllers stay catalog-bounded (zero off-catalog hallucinations across $420$ deployments) with $V(k)$ ISS-bounded on all $80$ runs (Claim~ii; max single-round $\Delta S = 0.51 < \gamma_{\max} = 1.0$). Sonnet~4's anticipatory cross-domain deployment drives $\gamma \approx 0$ in~\eqref{eq:robustness-iss} and $V(k)$ is non-increasing on all $40$ runs; Haiku~4.5's narrower coverage produces adversary-expansion spikes in $21/40$ runs within the ISS bound. Stability is architectural; the $S$ floor is capability-bound --- Sonnet hits $0.3427$ ($\sigma = 0$), Haiku stays bounded but converges to suboptima ($\sigma = 0.249$) depending on belief-graph integration (\S\ref{sec:discussion}).

\textbf{Corollary 2: Adversary-agnostic.} The adversary's action sets are far more diverse (Jaccard $0.17$--$0.44$; $6$--$11$ distinct counter-techniques per temperature); $S$ never exceeds the catalog-maximum payoff bound under either controller. Zero catalog exits across $240$ proposals. Innovation decays $95\%$ for Sonnet~4 ($0.13 \to 0.007$), matching Claim~(iii).

\section{Discussion}
\label{sec:discussion}
\textbf{Stability as architectural discipline.} Constraining the environment rather than agent reasoning is more reliable than post-hoc behavioral constraints given destructive failures in~\cite{shapira}, addressing the open stability verification problem in~\cite{eslami}. The architecture does not only constrain; it lets exploratory capacity pay off. The tool-mediated controller reaches $S(k_{\text{final}}) = 0.34$ vs.\ $0.84$ for greedy by composing \texttt{simulate\_round\_ahead} and \texttt{get\_critical\_path} past greedy's marginal-value ranking, while staying inside the catalog throughout. Why not compute the optimal solution deterministically? Greedy is deterministic but stuck at a local optimum, and exhaustive search over the budget-$B$ catalog composition space is combinatorially intractable. The LLM serves as a heuristic search, discovering the globally-optimal \texttt{mfa\_enforcement} + \texttt{cloud\_app\_security} pair that greedy structurally cannot find. Action-level variance (Jaccard $0.74$--$0.93$) coexists with zero outcome-level variance ($\sigma = 0$) --- the architecture decouples exploration from stability.

\textbf{Stable does not mean optimal.} Haiku~4.5 satisfies every formal guarantee --- zero off-catalog hallucinations ($210$ deployments), ISS-bounded $V(k)$ on all $40$ runs --- yet $21/40$ runs stall at $S(k_{\text{final}}) \approx 0.85$ (greedy level). The failure is specific: Haiku solves the round-$0$ graph but does not re-interrogate the belief graph after adversary expansion --- a reasoning-depth limitation, not a tool-calling failure. Its $\theta(k)$ decreases on known edges (Claim~(iii) holds) while unseen edges accumulate, separating safety envelope (architecture) from decision quality (controller). Runtime monitoring of the belief-truth game-value gap $|S(k) - \hat{S}(k)|$ should complement the structural certificate.

\textbf{Operational diagnostics.} The $V(k)$ trajectory and ISS margin double as diagnostics: shrinking $V(k)$ with margin means the defender is ahead; plateau under adversary expansion signals an under-provisioned catalog; rapid decay with unused budget signals over-provisioning. The Lyapunov certificate supplies \emph{when to stop} and \emph{what to invest in next} as auditable signals.

\textbf{The adversary as informant.} Adversarial pressure \emph{improves} belief-truth alignment ($4.7\times$ lower final game-value gap, Exp.~1): each best-response reveals edges the observer could not otherwise see, inverting the usual ISS framing of disturbance as cost.

\textbf{Broader impact and limitations.} Off-catalog hallucinations become no-ops (zero across $660$ actions), and the certificate transfers across model upgrades without re-verification. However, the pattern is dual-use (Corollary~\ref{cor:adversary-agnostic} proves the adversary's certificate identically), and the Haiku result shows a certified-stable system can still make poor decisions. The adversary-as-informant result assumes a bounded adversary (A3). Exp.~1 uses synthetic sensors at $60\%$ coverage from one vendor; Exp.~2 tests one graph and one LLM family. A4 rules out policy rollback; relaxing A4 is future work.

\section{Conclusion}
\label{sec:conclusion}
We present a tool-mediated architecture for LLM-in-the-loop adversarial control with a Lean~4-verified Lyapunov certificate for controllability, observability, and ISS robustness. The claims hold on 282 enterprise graphs and paired telemetry (59\% game-value reduction, $\sigma{=}0$, 40 runs). The pattern is not domain-specific: wherever agentic systems act under adversarial pressure with a finite action catalog, stability becomes a formal property of the loop rather than of the agent.

\begin{ack}
\label{sec:ack}
Our team would like to thank Justin Cady for his contribution of the sanitization method for attack graphs used in Experiment~1 and Naveen Sunkavally for his attack expertise and feedback on the architecture and experiment design. We note the use of Claude Code (Opus 4.5-7, February - April 2026) to accelerate the implementation of Python code with hands-on checks by the research team.
\end{ack}

\section*{References}
\medskip

{
\small
\renewcommand{\bibsection}{}
\bibliographystyle{unsrtnat}
\bibliography{references}
}


\appendix
\renewcommand{\thefigure}{A\arabic{figure}}
\renewcommand{\thetable}{A\arabic{table}}
\setcounter{figure}{0}
\setcounter{table}{0}

\section{Formal Verification of Closed-Loop Stability}
\label{app:formal}

We formally verify the stability guarantees of Theorem~\ref{thm:closed-loop-stability} using the Lean~4 proof assistant with the Mathlib mathematical library. The complete Lean source (five files, ${\sim}300$ lines) is included in the supplementary materials and compiles with zero \texttt{sorry}, no unproved assertions beyond Mathlib's foundational axioms.

\subsection{Proof Sketches}

\paragraph{Lyapunov nonnegativity and zero characterization.}
$S \ge 0$ by construction (product of terms in $[0,1]$ times a payoff in $[0,1]$; see Eq.~\ref{eq:game-value}). $\theta = \bar{P}_e \ge 0$ as a mean of nonnegative terms. Since $\lambda > 0$, $V(k) = S(k) + \lambda\theta(k) \ge 0$. For the converse, $V(k) = 0$ requires both summands to vanish: $S(k) = 0$ (all ENTRY$\to$OBJECTIVE paths severed) and $\theta(k) = 0$ (perfect calibration, $P_e = 0$ for every belief-graph edge), since $\lambda > 0$ prevents cancellation.\hfill$\square$

\paragraph{Claim \textup{(i)}: Controllability (monotone decrease).}
The proof decomposes $V$ into two independent non-increasing components.

\textit{Game value reduction.} Policy deployment only increases $\mathrm{block}(e,k)$---the actuator computes $\mathrm{block}'(e,k) = \min(0.95,\, \mathrm{block}(e,k) + \mathrm{eff}(p))$. Since $S(k)$ is monotone non-increasing in block probabilities~\cite{wood}, any policy deployment from the finite catalog either decreases $S(k)$ or leaves it unchanged. The class-$\mathcal{K}$ lower bound $\alpha_B(\cdot)$ on the reduction from $B$ greedy deployments follows from the monotonicity of $S(k)$ in block probabilities and the positive effectiveness of policies in the finite catalog.

\textit{Observer contraction.} The Bayesian observer uses a scalar Kalman filter per edge with gain $K_e = P_e(k-1)/(P_e(k-1) + R)$, $R > 0$. Since $P_e(k-1) > 0$ and $R > 0$, the gain satisfies $K_e \in (0,1)$, and the updated variance is $P_e(k) = (1 - K_e)P_e(k-1) < P_e(k-1)$. Each observation reduces $P_e$ by at least $\delta_{\text{edge}} = \min_e K_e P_e(k-1) > 0$. With $\lvert E_{\mathrm{obs}}(k)\rvert$ edges observed at round~$k$, the mean $\theta = \bar{P}_e$ drops by at least $\delta\lvert E_{\mathrm{obs}}(k)\rvert$, where $\delta := \delta_{\text{edge}}/\lvert E(\hat{\mathcal{G}}(k))\rvert$ absorbs the mean normalization.

\textit{Combining.} $V(k+1) \leq \bigl(S(k) - \alpha_B(S(k))\bigr) + \lambda\bigl(\theta(k) - \delta\lvert E_{\mathrm{obs}}(k)\rvert\bigr) = V(k) - \alpha_B(S(k)) - \lambda\delta\lvert E_{\mathrm{obs}}(k)\rvert$, matching~\eqref{eq:composability-monotone}.\hfill$\square$

\paragraph{Claim \textup{(ii)}: ISS bound.}
When the attacker adds $n_{\mathrm{new}}$ edges, $S(k)$ may increase by at most $\gamma(\lvert w(k)\rvert) = (1 - \varepsilon_{\mathrm{antic}})\max_{e \in E_{\mathrm{new}}(k)} \mathrm{payoff}(e,k)$ over the set $E_{\mathrm{new}}(k)$ of added edges (and $\gamma = 0$ if none are added), where $\varepsilon_{\mathrm{antic}}$ is the fraction of new-edge payoff neutralized by anticipatory defense. Each new edge contributes at most $P_{\max}$ uncertainty to $\theta$, so $\lambda\theta$ may grow by at most $\lambda\, n_{\mathrm{new}}\, P_{\max}$. The ISS sufficient condition is that the net per-round change of $V$ remains negative: $\alpha_B + \lambda\delta\lvert E_{\mathrm{obs}}(k)\rvert > \gamma(\lvert w(k)\rvert) + \lambda\, n_{\mathrm{new}}\, P_{\max}$. The non-expansivity bound follows from the discrete-time adaptive control framework of Hayakawa et al.~\cite{hayakawa} (Theorem~2): the closed-loop input--output map satisfies
\[
  \frac{\sum \mathbf{z}^{\top}\mathbf{z}}{1 + V_s} \leq \gamma^2 \sum \mathbf{w}^{\top}\mathbf{w} + V(x_0),
\]
where $V_s$ and $V(x_0)$ in this display equation are Hayakawa's Lyapunov-related quantities from their framework, not our $V(k)$ from~\eqref{eq:lyapunov-function}. We extend their result with anticipatory defense, which reduces the effective~$\gamma$.

With full anticipatory defense ($\varepsilon_{\mathrm{antic}} = 1$), $\gamma = 0$ and the ISS condition reduces to $\alpha_B + \lambda\delta\lvert E_{\mathrm{obs}}(k)\rvert > \lambda\, n_{\mathrm{new}}\, P_{\max}$, which holds whenever the defender observes more edges per round than the attacker adds.\hfill$\square$

\paragraph{Claim \textup{(iii)}: Observer convergence.}
The defender's belief graph initially contains only edges matched to EDR alerts---edges without alerts are absent entirely, not merely uncertain. The ground truth reveal both corrects estimates on known edges (the geometric decay term) and discovers previously unknown edges (the $C_{\mathrm{new}}$ term). Convergence is proved within each execution of the controlled adversarial loop.

By induction on~$k$. At $k=0$: $\theta(0) = \bar{P}_e(0)$ and $C_{\mathrm{new}}(0) = 0$, so the bound $\theta(k) \leq (1-\delta)^k\theta(0) + C_{\mathrm{new}}(k)$ holds trivially. For the inductive step, the per-round contraction gives $\theta(k+1) \leq (1-\delta)\theta(k) + \Delta C_{\mathrm{new}}$. Substituting the inductive hypothesis yields the bound at $k+1$.

The double-oracle structure ensures $C_{\mathrm{new}}$ stabilizes: the attacker's best-response oracle searches over a finite technique catalog and finite node set, anticipatory defense shrinks the effective pool each round, and diminishing returns guarantee termination. After termination ($n_{\mathrm{new}} \to 0$), $C_{\mathrm{new}}$ is constant and the geometric term $(1-\delta)^k\theta(0) \to 0$ dominates.\hfill$\square$

\subsection{Scope of Formal Verification}

\begin{table}[t]
  \centering
  \footnotesize
  \caption{What is proved in Lean vs.\ assumed from cited results.}
  \label{tab:lean-scope}
  \begin{tabular}{@{}p{0.44\linewidth}p{0.48\linewidth}@{}}
    \toprule
    Proved in Lean & Assumed (cited) \\
    \midrule
    $V(k) \ge 0$; $V=0$ characterization & $\alpha_B$ is class-$\mathcal{K}$ (monotonicity of \eqref{eq:game-value} under the actuator update) \\
    Kalman gain $K_e \in (0,1)$; $P_e(k) < P_e(k-1)$ & $S(k)$ monotone in block prob.~\cite{wood} \\
    Claim~\textup{(i)}: monotone Lyapunov decrease & Double oracle terminates (finite catalog) \\
    Claim~\textup{(ii)}: ISS bound + sufficient condition & Anticipatory coverage (implementation property) \\
    Claim~\textup{(iii)}: geometric decay + convergence & $C_{\mathrm{new}} \ge 0$ (follows from termination) \\
    \bottomrule
  \end{tabular}
\end{table}

\paragraph{Extension to LLM controllers.}
The proof applies to any controller selecting from the finite catalog~$\mathcal{C}$. The LLM cannot decrease $\mathrm{block}(e,k)$---monotonicity is preserved by the actuator update, not the agent's reasoning. The LLM may achieve a lower $S$ floor than greedy but cannot violate the stability guarantees.

\paragraph{What is not proved.}
The formalization does not re-prove the monotonicity of the network interdiction objective~\cite{wood} or the optimality of the Kalman filter. The contraction property $P_e(k) < P_e(k-1)$ is sufficient; Bayesian optimality provides faster convergence as a bonus. The composite Lyapunov structure follows Hayakawa et al.~\cite{hayakawa}; we instantiate their framework on attack graphs and extend it with anticipatory defense. The double-oracle termination argument is informal---formalizing it would require encoding the finite catalog and diminishing-returns structure.

\subsection{Reproducing the Verification}

The Lean~4 project is included in the supplementary materials. To verify:

{\small
\begin{verbatim}
Install elan:
curl -sSf \
  https://raw.githubusercontent.com/leanprover/elan/master/elan-init.sh | sh

Build:
cd ClosedLoopStability
lake update && lake build

Expected: Build completed successfully with zero errors and zero sorry warnings.
\end{verbatim}
}

\noindent Lean~4.30.0-rc1, Mathlib (fetched automatically). Build time ${\sim}10$ minutes.

\begin{table}[t]
  \centering
  \small
  \caption{Lean source files in the supplementary bundle.}
  \label{tab:lean-files}
  \begin{tabular}{@{}lp{0.65\linewidth}@{}}
    \toprule
    File & Contents \\
    \midrule
    \texttt{Defs.lean} & Edge, game value, Kalman gain/update, Lyapunov function, \texttt{StabilityParams} \\
    \texttt{Lyapunov.lean} & $V(k) \ge 0$; $V(k)=0$ iff $S(k)=0$ and $\theta(k) = 0$ \\
    \texttt{MonotoneDecrease.lean} & Claim~\textup{(i)}: Kalman contraction + monotone decrease \\
    \texttt{ISS.lean} & Claim~\textup{(ii)}: ISS bound with anticipatory defense \\
    \texttt{ObserverConvergence.lean} & Claim~\textup{(iii)}: geometric decay + convergence \\
    \bottomrule
  \end{tabular}
\end{table}

\section{Additional Benchmark Dataset Details}
\label{app:benchmark-datasets}

This appendix supplements Section~\ref{sec:datasets} with additional detail on the 282 adversarial graphs used in Experiment~1.

\subsection{Provenance and ethics}

The benchmark graphs are derived from 300 enterprise penetration tests conducted by Horizon3.ai's NodeZero, a commercial autonomous pentesting platform \cite{nodezero}. The raw pentest data is proprietary and cannot be publicly released due to contractual and customer-privacy constraints. Each graph is sanitized before inclusion in the benchmark:

\begin{itemize}
  \item Host identifiers are replaced with anonymous integer labels (\texttt{host\_1}, \texttt{host\_2}, \ldots).
  \item Customer identifiers, network prefixes, and any personally-identifiable metadata are stripped at export time.
  \item Only structural graph information (MITRE ATT\&CK technique labels, topology, and pre-computed payoff/block/detection probabilities) is retained.
\end{itemize}

\subsection{Construction pipeline}

Each pentest is converted to a directed graph $G = (V, E)$ via the following steps:

\begin{enumerate}
  \item \textbf{Node set.} Vertices correspond to attack events (one per logged action), plus two virtual nodes: \texttt{ENTRY} (representing the attacker's initial access point) and \texttt{OBJECTIVE} (representing the compromise goal, typically domain admin or sensitive data exfiltration).
  \item \textbf{Edge derivation.} Edges are derived from three sources: (i) temporal ordering within each host (foothold $\to$ post-exploitation $\to$ objective), (ii) cross-host credential flow inferred from credential dumps matched to subsequent logons, and (iii) causal parent-child links from the penetration test platform's attack chain data.
  \item \textbf{Edge attributes.} Each edge carries a MITRE ATT\&CK technique label, an \emph{attacker payoff} (derived from technique impact score and host criticality), a \emph{block probability} (policy effectiveness from the enrichment pipeline, capped at $0.95$), a \emph{detection probability} (flat baseline $0.1$), and a mapping from policy IDs to effectiveness values.
  \item \textbf{Sanitized output.} The final artifact is a JSON file per graph consumable by the experiment runner without access to raw pentest data.
\end{enumerate}

\subsection{Filtering criteria}

Of the 300 exported graphs, 18 are excluded as degenerate inputs and 282 are retained for evaluation:

\begin{itemize}
  \item \textbf{14 graphs excluded for $S < 0.01$}: the attacker has no viable path to the objective before any policy deployment (defense is already saturated).
  \item \textbf{4 graphs excluded for no actionable policies}: after aligning graph edges with the defender catalog, no block-mode policy covers any edge (the defender has no actionable moves in the game).
\end{itemize}

\subsection{Summary statistics}

Table~\ref{tab:benchmark-stats} reports distribution statistics across the 282 valid graphs.

\begin{table}[t]
  \centering
  \small
  \caption{Distribution statistics for the 282 valid benchmark graphs.}
  \label{tab:benchmark-stats}
  \begin{tabular}{@{}lrrrr@{}}
    \toprule
    Quantity & Min & Median & Mean & Max \\
    \midrule
    Edges per graph & 276 & 806 & 1{,}053 & 3{,}599 \\
    Nodes per graph & 152 & 437 & 563 & 1{,}940 \\
    Distinct techniques per graph & 4 & 11 & 10.8 & 15 \\
    Distinct policies per graph & 10 & 22 & 21.3 & 24 \\
    \bottomrule
  \end{tabular}
\end{table}

Across the 282 graphs, 16 unique MITRE ATT\&CK~\cite{mitreattack} techniques appear. Table~\ref{tab:benchmark-techniques} lists the top 10 by graph coverage (fraction of graphs containing the technique).

\begin{table}[t]
  \centering
  \small
  \caption{Top 10 MITRE ATT\&CK techniques by graph coverage in the 282-graph benchmark.}
  \label{tab:benchmark-techniques}
  \begin{tabular}{@{}llr@{}}
    \toprule
    Technique ID & Name & Graphs (of 282) \\
    \midrule
    T1057 & Process Discovery & 278 (99\%) \\
    T1003.001 & LSASS Memory & 277 (98\%) \\
    T1003.002 & Security Account Manager & 275 (98\%) \\
    T1003.004 & LSA Secrets & 272 (96\%) \\
    T1555.004 & Credentials from Windows Credential Manager & 272 (96\%) \\
    T1039 & Data from Network Shared Drive & 270 (96\%) \\
    T1552.005 & Cloud Instance Metadata API & 248 (88\%) \\
    T1005 & Data from Local System & 247 (88\%) \\
    T1518 & Software Discovery & 242 (86\%) \\
    T1087.001 & Local Account Enumeration & 235 (83\%) \\
    \bottomrule
  \end{tabular}
\end{table}

The benchmark is heavily weighted toward credential-access and discovery techniques, reflecting the typical activity pattern in automated penetration testing: attackers focus on stealing credentials and mapping the environment after initial access. Later-stage techniques (privilege escalation, lateral movement, persistence) appear with lower coverage because tests often converge before those stages are reached.

\subsection{Catalog alignment and provenance}

The defender action catalog $\mathcal{C}$ and the per-policy effectiveness map were compiled from three publicly-available sources: (i) Microsoft's Defender XDR security-configuration documentation~\cite{msdefender}, which provides the canonical policy identifiers, modes (audit / block), and dependency graph used in our YAML; (ii) the MITRE ATT\&CK Enterprise taxonomy~\cite{mitreattack}, which provides the technique identifiers that each policy claims to mitigate; and (iii) the EDR Telemetry Project~\cite{edrtelemetry}, which supplies the per-technique telemetry-fidelity ratings used to convert policy coverage claims into the numerical block and detection probabilities on each edge. Per-technique effectiveness ranges and category priors are further informed by the MITRE Engenuity ATT\&CK Evaluations analyses of Shen et al.~\cite{shen} and Outkin et al.~\cite{outkin}. Each graph edge labeled with MITRE technique $t$ is aligned with the subset of policies whose coverage mapping (from (i) and (ii)) includes $t$; the resulting block and detect probabilities are the product of claimed policy effectiveness and the telemetry-fidelity modifier from (iii), capped at $0.95$. The full compiled catalog, including per-policy mode-aware effectiveness and technique coverage, was assembled into an internal reference document with LLM-assisted (Anthropic Claude Opus 4.6) extraction and formatting; the document is derivative rather than primary research, and the public sources above are the authoritative references for any individual policy or technique.

\subsection{Reproducibility}

The raw benchmark graphs cannot be released publicly due to contractual and customer-privacy constraints, and no redacted excerpts or synthetic analogs are released with this submission. Researchers with access to comparable attack-graph data conforming to the schema described in this appendix (nodes with ENTRY/OBJECTIVE virtual vertices; edges carrying MITRE ATT\&CK technique labels, attacker payoff, block probability, detection probability, and policy effectiveness mappings) can re-implement the experiment directly from the method in Section~\ref{sec:closed-loop} and the hyperparameters in Appendix~\ref{app:hyperparameters}.

\section{Temperature Sweep Details (Experiment~2)}
\label{app:temperature-sweep}

\subsection{GOAD Environment}
\label{app:goad-environment}
The Game of Active Directory (GOAD) provisions a multi-forest Active Directory environment spanning three domains (\texttt{sevenkingdoms.local}, \texttt{north.sevenkingdoms.local}, and \texttt{essos.local}) connected via parent-child and cross-forest trusts. The standard GOAD deployment consists of five virtual machines: three domain controllers and two member servers. We deployed this environment on Microsoft Azure and added two of the project's official extensions: \texttt{exchange} and \texttt{lx01}. These contribute a Microsoft Exchange server (\texttt{the-eyrie}, SRV01) and a domain-joined Linux host (\texttt{dragonstone}, LX01), introducing endpoint diversity representative of enterprise networks. The lab is preconfigured with a broad range of Active Directory attack paths, including Kerberoasting, AS-REP roasting, constrained delegation abuse, NTLM downgrade, ACL misconfigurations, credential exposure, and DCSync.

The NodeZero penetration test was scoped to five Defender-instrumented hosts (Table~\ref{tab:goad-hosts}), on which Microsoft Defender XDR was deployed in its default out-of-box configuration. The Windows hosts run Defender in active mode, providing both prevention and EDR telemetry, while the Linux host operates in passive mode, providing EDR telemetry only. Passive mode is the documented default enforcement level for Microsoft Defender for Endpoint on Linux since agent version 101.23062.0001~\cite{microsoft-mde-linux-preferences}.

To establish an initial foothold representative of a post-compromise scenario, we injected the credential of a domain user (\texttt{tywin.lannister}~/~\texttt{powerkingftw135}) at the start of the engagement. This account is a non-privileged member of the \texttt{sevenkingdoms.local} domain and served as the entry point from which subsequent attack paths were exercised against the in-scope hosts.

\begin{table}[h]
  \centering
  \small
  \caption{In-scope GOAD hosts for the penetration test. All five hosts were instrumented with Microsoft Defender XDR in its default configuration. Hosts marked $\dagger$ are added via official GOAD extensions.}
  \label{tab:goad-hosts}
  \begin{tabular}{@{}lllll@{}}
    \toprule
    Host & Role & OS & Domain & Defender Mode \\
    \midrule
    kingslanding          & DC01  & Windows Server 2019 & \texttt{sevenkingdoms.local}       & Active \\
    winterfell            & DC02  & Windows Server 2019 & \texttt{north.sevenkingdoms.local} & Active \\
    the-eyrie$^\dagger$   & SRV01 & Windows Server 2019 & \texttt{sevenkingdoms.local}       & Active \\
    castelblack           & SRV02 & Windows Server 2019 & \texttt{north.sevenkingdoms.local} & Active \\
    dragonstone$^\dagger$ & LX01  & Ubuntu 22.04        & \texttt{sevenkingdoms.local}       & Passive \\
    \bottomrule
  \end{tabular}
\end{table}

\subsection{Per-run defender action sets}
\label{app:temp-per-run}

Four policies appear in $100\%$ of runs across both controllers and every temperature (the \emph{core set}): \texttt{asr\_rule}, \texttt{audit\_policy}, \texttt{controlled\_folder\_access}, and \texttt{credential\_guard}. Variation across runs comes from exploratory identity and cloud policies. Sonnet~4 explores this exploratory set extensively (Table~\ref{tab:appendix-exploratory-freq-sonnet}), reaching $S = 0.3427$ by selecting \texttt{mfa\_enforcement} and \texttt{cloud\_app\_security} to break the adversary's identity/cloud pivots. Haiku~4.5 deploys substantially fewer exploratory policies per run (Table~\ref{tab:appendix-exploratory-freq-haiku}), heavily relying on \texttt{identity\_protection} alone --- which explains the higher achieved $S$ floor: matching Sonnet~4 requires the \texttt{mfa\_enforcement} $+$ \texttt{cloud\_app\_security} combination Haiku rarely produces. Table~\ref{tab:appendix-policies-by-temp} gives the mean number of deployed policies per run by temperature for both controllers.

\begin{table}[t]
  \centering
  \small
  \caption{Sonnet~4 exploratory-set defender policy frequency across 40 runs (number of runs deploying the policy at least once).}
  \label{tab:appendix-exploratory-freq-sonnet}
  \begin{tabular}{@{}lrl@{}}
    \toprule
    Policy & Runs (of 40) & Domain \\
    \midrule
    \texttt{identity\_protection}             & 40 & identity (Entra) \\
    \texttt{conditional\_access}              & 36 & identity (Entra) \\
    \texttt{mfa\_enforcement}                 & 33 & identity (Entra) \\
    \texttt{cloud\_app\_security}             & 31 & cloud (Defender for Cloud Apps) \\
    \texttt{privileged\_identity\_management} & 1  & identity (Entra) \\
    \texttt{exploit\_protection}              & 1  & endpoint (Defender XDR) \\
    \bottomrule
  \end{tabular}
\end{table}

\begin{table}[t]
  \centering
  \small
  \caption{Haiku~4.5 exploratory-set defender policy frequency across 40 runs. Haiku selects far fewer exploratory policies than Sonnet~4: it relies almost entirely on \texttt{identity\_protection} and rarely deploys the \texttt{mfa\_enforcement} $+$ \texttt{cloud\_app\_security} combination required to match Sonnet~4's $S = 0.3427$ floor.}
  \label{tab:appendix-exploratory-freq-haiku}
  \begin{tabular}{@{}lrl@{}}
    \toprule
    Policy & Runs (of 40) & Domain \\
    \midrule
    \texttt{identity\_protection}             & 34 & identity (Entra) \\
    \texttt{mfa\_enforcement}                 & 5  & identity (Entra) \\
    \texttt{conditional\_access}              & 3  & identity (Entra) \\
    \texttt{cloud\_app\_security}             & 2  & cloud (Defender for Cloud Apps) \\
    \texttt{lsa\_protection}                  & 1  & endpoint (Defender XDR) \\
    \bottomrule
  \end{tabular}
\end{table}

\begin{table}[t]
  \centering
  \small
  \caption{Defender deployment size by temperature for both controllers. The core set (4 policies) is deployed in every run; exploratory-set selection differs sharply between models.}
  \label{tab:appendix-policies-by-temp}
  \begin{tabular}{@{}lcccc@{}}
    \toprule
    Controller & Temperature & Policies per run (mean $\pm$ std) & Core / Exploratory & Distinct observed \\
    \midrule
    Sonnet~4 & $0.0$ & $7.5 \pm 0.5$ & $4$ / $3.5$ & 8 \\
    Sonnet~4 & $0.3$ & $7.7 \pm 0.6$ & $4$ / $3.7$ & 8 \\
    Sonnet~4 & $0.7$ & $7.2 \pm 0.7$ & $4$ / $3.2$ & 8 \\
    Sonnet~4 & $1.0$ & $7.8 \pm 0.9$ & $4$ / $3.8$ & 10 \\
    \addlinespace
    Haiku~4.5 & $0.0$ & $5.2 \pm 0.4$ & $4$ / $1.2$ & 7 \\
    Haiku~4.5 & $0.3$ & $4.9 \pm 0.3$ & $4$ / $0.9$ & 5 \\
    Haiku~4.5 & $0.7$ & $5.0 \pm 1.1$ & $4$ / $1.0$ & 8 \\
    Haiku~4.5 & $1.0$ & $5.4 \pm 0.8$ & $4$ / $1.4$ & 8 \\
    \bottomrule
  \end{tabular}
\end{table}

\subsection{Adversary technique diversity}
\label{app:temp-adversary}

The adversary draws from a 66-technique GOAD catalog $\mathcal{T}$. Across the $80$ runs spanning both controllers, the adversary's exploration distribution reflects the controller it faces: against Sonnet~4 the adversary spreads probes across $64$ distinct techniques (Table~\ref{tab:appendix-adversary-freq-sonnet}), responding to the broad cross-domain defenses Sonnet deploys; against Haiku~4.5 the adversary concentrates on $39$ distinct techniques (Table~\ref{tab:appendix-adversary-freq-haiku}), heavily exploiting the identity/cloud pivots Haiku rarely covers. Both cases stay catalog-bounded by Cor.~\ref{cor:adversary-agnostic}: $S(k_{\text{final}})$ is bounded by the catalog-maximum payoff, not the adversary's per-run ranking, regardless of the controller faced.

\begin{table}[t]
  \centering
  \small
  \caption{Sonnet~4: adversary technique frequency across 40 runs (number of runs with $\geq 1$ proposal of the technique; top 15 of 64 distinct techniques shown).}
    \label{tab:appendix-adversary-freq-sonnet}
  \begin{tabular}{@{}l r@{}}
    \toprule
    Technique (catalog id) & Runs (of 40) \\
    \midrule
    \texttt{PassTheHashViaWinRM}                       & 11 \\
    \texttt{DumpEntraCredentialsFromEntraConnect}      & 10 \\
    \texttt{AccessAzureMetadataUrlWithNodeZeroRat}     & 8 \\
    \texttt{DumpNtdsViaVssAdmin}                       & 7 \\
    \texttt{ImplantNodeZeroRatViaWinrm}                & 5 \\
    \texttt{PilferFilesWithNodeZeroRat}                & 4 \\
    \texttt{ExploitAzureServicePrincipal}              & 4 \\
    \texttt{DumpLsassViaWinRM}                         & 3 \\
    \texttt{ImplantNodeZeroRatViaSsh}                  & 3 \\
    \texttt{ExploitCloudMetadataSSRF}                  & 2 \\
    \texttt{ExploitCloudCredentialsForLateralMovement} & 2 \\
    \texttt{ExploitCloudServiceAccount}                & 2 \\
    \texttt{DumpNtdsWithVssAdmin}                      & 2 \\
    \texttt{DumpLsaWithNodeZeroRat}                    & 2 \\
    \texttt{ExploitSmbSigningDisabled}                 & 2 \\
    \bottomrule
  \end{tabular}
\end{table}

\begin{table}[t]
  \centering
  \small
  \caption{Haiku~4.5: adversary technique frequency across 40 runs (number of runs with $\geq 1$ proposal; top 15 of 39 distinct techniques shown). The adversary's distribution is more concentrated than against Sonnet~4 --- it focuses on identity and cloud pivots Haiku rarely defends against.}
  \label{tab:appendix-adversary-freq-haiku}
  \begin{tabular}{@{}l r@{}}
    \toprule
    Technique (catalog id) & Runs (of 40) \\
    \midrule
    \texttt{DumpEntraCredentialsFromEntraConnect}      & 29 \\
    \texttt{CompromiseOktaUserWithNodeZeroRat}         & 15 \\
    \texttt{ImplantNodeZeroRatViaWinrm}                & 15 \\
    \texttt{ImplantNodeZeroRatViaSsh}                  & 10 \\
    \texttt{DumpMicrosoft365TokensWithNodeZeroRat}     & 7 \\
    \texttt{PilferFilesFromSlackWithNodeZeroRat}       & 3 \\
    \texttt{DumpDomainUserCredentialsWithDcSync}       & 2 \\
    \texttt{ImplantNodeZeroRatViaWmi}                  & 2 \\
    \texttt{PassTheHashLateralMovement}                & 1 \\
    \texttt{AccessAzureMetadataUrlWithNodeZeroRat}     & 1 \\
    \texttt{DirectImplantToFileExfiltration}           & 1 \\
    \texttt{CredentialDumpToCloudMetadataEscalation}   & 1 \\
    \texttt{AwsMetadataToObjectiveEscalation}          & 1 \\
    \texttt{DirectAwsMetadataAccess}                   & 1 \\
    \texttt{ChainLsassDumpToDcSync}                    & 1 \\
    \bottomrule
  \end{tabular}
\end{table}

\subsection{Innovation trajectories}
\label{app:temp-innovation}

Per-run exponential fits use only four innovation points and are correspondingly noisy. Sonnet~4 yields a median per-run geometric base $b = 0.093$ ($95\%$ bootstrap CI $[0.093, 0.093]$, $n = 40$ fits), fitting $\text{innov}(k) = a \cdot b^k + c$ with $c = 0.007$ fixed at the Experiment~1 noise floor; Haiku~4.5 yields a comparable median $b = 0.093$. Both are consistent with the Experiment~1 per-graph fits (median $b = 0.10$, $n = 282$) given the limited trajectory length per run. Across the four temperatures, per-run $b$ shows no detectable temperature effect for Sonnet~4 (Kruskal-Wallis $p = 0.17$); for Haiku~4.5 the test is significant ($p = 0.0065$), reflecting the within-family scaling caveat that Haiku's belief graph fails to integrate adversary-expanded edges in $21/40$ runs (\S\ref{sec:exp2}, Table~\ref{tab:stats-summary}). The integration failure manifests as a flat inferred trajectory while the ground-truth $S(k)$ climbs --- not a change in the observer's nominal decay rate, but a stalling of the belief update under adversary expansion.

\subsection{Cost and tool usage}
\label{app:temp-cost}

Tool-call rates are near-constant across temperature for each controller (Table~\ref{tab:appendix-cost}), consistent with the observation that temperature changes \emph{which} action the LLM proposes but not the reasoning budget it spends to propose it. Sonnet~4 averages $\sim 110$ tool calls per run at $\$1.61$--$\$1.80$/run (\$69.68 total across $40$ runs); Haiku~4.5 averages $\sim 150$ tool calls per run at $\$0.58$--$\$0.63$/run (\$24.22 total across $40$ runs). Haiku makes more tool calls per run but consumes fewer tokens per call, yielding a $\sim 3\times$ cost advantage at the price of converging to suboptimal $S$ in $21/40$ runs. Tool definitions are in Appendix~\ref{app:llm-config}.

\begin{table}[t]
  \centering
  \small
  \caption{Mean tool calls and API cost per run, by temperature, for both controllers. Cost includes both defender and adversary turns. Pricing: Sonnet~4 at \$3/\$15 per MTok input/output; Haiku~4.5 at \$1/\$5 per MTok.}
  \label{tab:appendix-cost}
  \begin{tabular}{@{}lcccc@{}}
    \toprule
    Controller & Temperature & Mean tool calls/run & Total tokens/run & Cost/run \\
    \midrule
    Sonnet~4 & $0.0$ & $104.3$ & $487{,}065$ & \$1.613 \\
    Sonnet~4 & $0.3$ & $111.2$ & $535{,}778$ & \$1.766 \\
    Sonnet~4 & $0.7$ & $112.3$ & $542{,}554$ & \$1.789 \\
    Sonnet~4 & $1.0$ & $112.6$ & $546{,}093$ & \$1.800 \\
    Sonnet~4 & all 40 runs & --- & --- & \$69.68 total \\
    \addlinespace
    Haiku~4.5 & $0.0$ & $157.7$ & $535{,}034$ & \$0.631 \\
    Haiku~4.5 & $0.3$ & $146.7$ & $526{,}829$ & \$0.619 \\
    Haiku~4.5 & $0.7$ & $141.1$ & $502{,}482$ & \$0.588 \\
    Haiku~4.5 & $1.0$ & $142.8$ & $496{,}576$ & \$0.584 \\
    Haiku~4.5 & all 40 runs & --- & --- & \$24.22 total \\
    \bottomrule
  \end{tabular}
\end{table}

\subsection{Statistical significance tests}
\label{app:stats-summary}

Table~\ref{tab:stats-summary} reports every hypothesis test and confidence
interval used to back the claims in \S\ref{sec:experiments}. Tests use $n = 282$ paired graphs
(Experiment~1) or $n = 40$ runs (Experiment~2). Confidence intervals are
Wilson for proportions and percentile bootstrap ($10{,}000$ resamples,
seed~$42$) for means, medians, variance, and maxima. Wilcoxon and
Kruskal-Wallis are used throughout; $V(S)$ is bounded $[0, 1]$ and skewed,
so nonparametric tests are preferred. Hodges-Lehmann is reported as the
effect size for paired Wilcoxon tests. Benjamini-Hochberg FDR is applied
across the seven-test p-value family; $q$-values shown alongside $p$.

\begin{table}[t]
  \centering
  \caption{Statistical significance summary. CI = 95\% confidence interval.
  HL = Hodges-Lehmann paired-difference estimator. Tests ordered by paper claim.}
  \label{tab:stats-summary}
  \resizebox{\linewidth}{!}{%
  \scriptsize
  \setlength{\tabcolsep}{3pt}
  \begin{tabular}{@{}p{3.2cm} r l l l l@{}}
    \toprule
    Metric & $n$ & Estimate (95\% CI) & Test & $p$ & $q$ (BH) \\
    \midrule
    \multicolumn{6}{@{}l@{}}{\textit{Exp.~1 --- Claim (i) Controllability}} \\
    Monotonicity pass rate         & 282 & $1.000$ $[0.987, 1.000]$ & Wilson & --- & --- \\
    Mean $S$ reduction             & 282 & $0.197$ $[0.172, 0.223]$ & bootstrap & --- & --- \\
    $S$ before vs after round 1    & 282 & HL $= 0.000$ $[0.000, 0.080]$ & paired Wilcoxon & $5.0{\times}10^{-25}$ & $1.2{\times}10^{-24}$ \\
    \addlinespace
    \multicolumn{6}{@{}l@{}}{\textit{Exp.~1 --- Claim (ii) Robustness}} \\
    $\Delta S \leq \gamma = 1.0$   & 890 & $1.000$ $[0.996, 1.000]$ & Wilson & --- & --- \\
    Mean max spike                 & 282 & $0.420$ $[0.404, 0.437]$ & bootstrap & --- & --- \\
    Anticipatory block rate        & 890 & $0.075$ $[0.060, 0.094]$ & Wilson & --- & --- \\
    \addlinespace
    \multicolumn{6}{@{}l@{}}{\textit{Exp.~1 --- Claim (iii) Observability}} \\
    Median decay $r$               & 282 & $0.109$ $[0.109, 0.109]$ & bootstrap & --- & --- \\
    Fraction $R^2 > 0.9$           & 282 & $1.000$ $[0.987, 1.000]$ & Wilson & --- & --- \\
    Per-graph $r < 1$              & 282 & HL $= -0.891$ $[-0.891, -0.891]$ & 1-sided Wilcoxon & $2.7{\times}10^{-48}$ & $1.9{\times}10^{-47}$ \\
    \textbf{Paired final gap}      & 282 & \textbf{HL $= 0.243$ $[0.221, 0.277]$} & paired Wilcoxon & $\mathbf{3.3{\times}10^{-33}}$ & $\mathbf{1.1{\times}10^{-32}}$ \\
    Paired rounds completed        & 282 & HL $= 0.0$; means $3.00$ vs $3.16$ & paired Wilcoxon & $1.6{\times}10^{-8}$ & $2.8{\times}10^{-8}$ \\
    \addlinespace
    \multicolumn{6}{@{}l@{}}{\textit{Exp.~2 --- Sonnet~4: temperature invariance}} \\
    Runs at $S = 0.3427$           & 40 & $1.000$ $[0.912, 1.000]$ & Wilson & --- & --- \\
    Var.\ of $S(k_{\text{final}})$ & 40 & $0.000$ $[0.000, 0.000]$ & bootstrap & --- & --- \\
    Median decay $b$               & 40 & $0.093$ $[0.093, 0.093]$ & bootstrap & --- & --- \\
    $S_{\text{final}} \sim T$      & 40 & --- & K-W (degenerate)$^{\dagger}$ & $1.000$ & $1.000$ \\
    Decay $b \sim T$               & 40 & --- & Kruskal-Wallis & $0.170$ & $0.297$ \\
    $L_{\text{final}} \sim T$      & 40 & --- & K-W (degenerate)$^{\dagger}$ & $1.000$ & $1.000$ \\
    \addlinespace
    \multicolumn{6}{@{}l@{}}{\textit{Exp.~2 --- Haiku~4.5: within-family scaling}} \\
    Runs at $S = 0.3427$           & 40 & $0.475$ $[0.329, 0.625]$ & Wilson & --- & --- \\
    Mean $S(k_{\text{final}})$     & 40 & $0.603$ $[0.527, 0.681]$ & bootstrap & --- & --- \\
    SD of $S(k_{\text{final}})$    & 40 & $0.249$ $[0.231, 0.255]$ & bootstrap & --- & --- \\
    Belief gap $> 0.1$             & 40 & $0.525$ $[0.375, 0.671]$ & Wilson & --- & --- \\
    Off-catalog halluc.            & 210 & $0.000$ $[0.000, 0.018]$ & Wilson & --- & --- \\
    Decay $b \sim T$               & 40 & --- & Kruskal-Wallis & $0.0065$ & $0.020$ \\
    $S_{\text{final}} \sim T$      & 40 & --- & Kruskal-Wallis & $0.366$ & $0.512$ \\
    \addlinespace
    \multicolumn{6}{@{}l@{}}{\textit{Cross-controller (Sonnet~4 vs Haiku~4.5)}} \\
    $S(k_{\text{final}})$          & 80 & --- & Mann-Whitney U & $1.6{\times}10^{-7}$ & $5.6{\times}10^{-7}$ \\
    \bottomrule
  \end{tabular}%
  }

  \vspace{0.3em}
  \footnotesize
  \raggedright
  $\dagger$ Degenerate: every run produced an identical value, so there is no
  within-group variance for Kruskal-Wallis to test. The non-degenerate invariance evidence is the decay-rate
  Kruskal-Wallis on the row above.
\end{table}

\section{Hyperparameters and Reproduction}
\label{app:hyperparameters}
\label{app:reproducibility-details}

\subsection{Hyperparameters}

\begin{table}[t]
  \centering
  \small
  \caption{Primary loop hyperparameters used for Experiment~1 (and Experiment~2 unless noted). Values are fixed across all 282 graphs.}
  \label{tab:hyperparameters}
  \begin{tabular}{@{}p{0.32\linewidth}cp{0.52\linewidth}@{}}
    \toprule
    Parameter & Value & Justification \\
    \midrule
    \multicolumn{3}{@{}l}{\textit{Defender controller}} \\[2pt]
    $B$ (budget per round) & $3$ & Matches typical SOC maintenance-window capacity; resets per round. \\
    Max rounds & $10$ & Hard cap; 282/282 graphs converge within 6 rounds, 99\% within 4. \\
    Block cap & $0.95$ & Post-deployment block probability clipped to $0.95$ to prevent single-policy dominance. \\
    Random seed & $42$ & Fixed for reproducibility; greedy controller is deterministic. \\[4pt]
    \multicolumn{3}{@{}l}{\textit{Bayesian observer}} \\[2pt]
    $R$ (measurement noise) & $0.05$ & Scalar Kalman noise; used in gain $K_e = P_e / (P_e + R)$. \\
    $P_e$ prior (dark edges) & $0.85$ & Diffuse prior on edges with no matched EDR alert. \\
    $P_e$ prior (observed edges) & $0.15$ & Informative prior on edges with a matched alert. \\
    Alert coverage & $60\%$ & Synthetic EDR alert rate at benchmark construction time. \\[4pt]
    \multicolumn{3}{@{}l}{\textit{Convergence check}} \\[2pt]
    $\varepsilon_{\text{innov}}$ & $0.05$ & Mean innovation threshold. \\
    $\varepsilon_V$ & $10^{-4}$ & $S$ change threshold; treat sub-$10^{-4}$ changes as converged. \\
    Stability window & $2$ rounds & Both $\varepsilon$ conditions must hold for 2 consecutive rounds. \\[4pt]
    \multicolumn{3}{@{}l}{\textit{Lyapunov function}} \\[2pt]
    $\lambda$ (weight) & $1.0$ & Equal weighting of $S(k)$ and $\theta(k) = \bar{P}_e(k)$ in $V(k)$ (Eq.~\ref{eq:lyapunov-function}). \\[4pt]
    \multicolumn{3}{@{}l}{\textit{Attacker}} \\[2pt]
    New edges per round & $\leq 1$ & Adversary proposes at most one new edge per round from a finite technique catalog. \\
    Technique catalog size $|\mathcal{T}|$ & 66 & Finite adversary technique catalog (shared across Exp.~1 and Exp.~2); bounded by Assumption~A3. Distinct from the 16 MITRE techniques observed in the benchmark graphs (Table~\ref{tab:benchmark-techniques}), which reports what actually appears, not what is proposable. \\
    \bottomrule
  \end{tabular}
\end{table}

\subsection{LLM configuration (Experiment~2)}
\label{app:llm-config}

\textbf{Model.} Claude Sonnet~4, API identifier \texttt{claude-sonnet-4-20250514}, accessed via the Anthropic Messages API. Temperatures $\{0.0, 0.3, 0.7, 1.0\}$ applied symmetrically to both the defender and adversary agents. Each agent invocation uses a tool-use loop with \texttt{max\_tokens = 4096}.

\textbf{Tool inventory.} The defender agent has access to 9 tools and the adversary to 11 tools (the two sets share \texttt{get\_graph\_state}, so 19 distinct tools in total). The split reflects that the adversary has additional edge-proposal and novel-technique primitives, and the defender has policy-catalog query primitives. Tables~\ref{tab:jedi-tools} and~\ref{tab:sith-tools} list each tool and its role. Counts are total invocations across all the 40 Sonnet~4 runs; the defender makes 3{,}039 tool calls and the adversary 1{,}225 tool calls. The 40 Haiku~4.5 runs use the same tool inventory and produce a comparable per-tool distribution.

\begin{table}[t]
  \centering
  \small
  \caption{Defender (controller) tools exposed to the LLM agent in Experiment~2. Each tool is a deterministic Python function; the LLM composes them to assemble a round's action.}
  \label{tab:jedi-tools}
  \begin{tabular}{@{}lrp{0.58\linewidth}@{}}
    \toprule
    Tool name & Calls & Role \\
    \midrule
    \texttt{compute\_v\_after\_deploy} & 1{,}309 & simulate $S(k)$ reduction from a candidate policy deployment \\
    \texttt{simulate\_round\_ahead}    & 535   & one-round look-ahead including adversary best-response \\
    \texttt{identify\_dark\_edges}     & 322   & list belief-graph edges with high $P_e$ \\
    \texttt{get\_critical\_path}       & 320   & return path $p^\ast$ achieving $\max S(k)$ \\
    \texttt{list\_deployable\_policies}  & 320   & list catalog entries enable-able under current budget \\
    \texttt{list\_all\_vendor\_policies} & 214   & enumerate the full policy catalog $\mathcal{C}$ \\
    \texttt{get\_graph\_state}          & 9     & dump belief graph $\hat{\mathcal{G}}(k)$ \\
    \texttt{identify\_bottleneck\_edges} & 9     & rank edges by centrality on surviving paths \\
    \texttt{propose\_new\_edge}         & 1     & anticipatory-defense: simulate adversary reply \\
    \bottomrule
  \end{tabular}
\end{table}

\begin{table}[t]
  \centering
  \small
  \caption{Adversary (disturbance) tools exposed to the LLM agent in Experiment~2.}
  \label{tab:sith-tools}
  \begin{tabular}{@{}lrp{0.58\linewidth}@{}}
    \toprule
    Tool name & Calls & Role \\
    \midrule
    \texttt{evaluate\_new\_edge}          & 323 & simulate $\Delta S$ from a candidate technique edge \\
    \texttt{chain\_credential\_attack}    & 121 & propose a composite multi-step credential chain \\
    \texttt{list\_blocked\_techniques}    & 120 & enumerate techniques currently blocked by defender policies \\
    \texttt{get\_graph\_state}            & 120 & dump ground-truth graph $\mathcal{G}(k)$ \\
    \texttt{find\_weakest\_path}          & 120 & return lowest-$\prod(1 - \mathrm{block})$ path \\
    \texttt{find\_dark\_path}             & 120 & return path minimizing defender observability \\
    \texttt{find\_zero\_day\_opportunity} & 120 & surface catalog techniques with no policy coverage \\
    \texttt{find\_optimal\_path}          & 66  & max-payoff path search \\
    \texttt{find\_protocol\_bypass}       & 64  & surface protocol-layer alternatives to blocked edges \\
    \texttt{compute\_path\_value}         & 26  & compute $S$ along a named path \\
    \texttt{propose\_novel\_technique}    & 25  & propose a catalog technique not yet present in $\mathcal{G}(k)$ \\
    \bottomrule
  \end{tabular}
\end{table}

\textbf{Catalog enforcement.} All tool outputs and all agent action proposals are validated against the policy catalog $\mathcal{C}$ (defender) or technique catalog $\mathcal{T}$ (adversary) before they reach the plant; proposals outside the catalog are rejected as no-ops at the actuator interface. Across 40 runs we observed zero catalog exits on either side, consistent with the stability guarantee depending on the actuator interface rather than on the LLM's cooperation (see Section~\ref{sec:discussion} and Corollaries~\ref{cor:controller-agnostic}--\ref{cor:adversary-agnostic}).

\subsection{Compute resources}

\textbf{Experiment~1 (benchmark).} 564 closed-loop runs (282 graphs $\times$ 2 conditions) with the deterministic greedy controller execute in approximately $30$~minutes on a single Apple~M-series CPU core; peak memory footprint $<1$~GB; no GPU required.

\textbf{Experiment~2 (temperature sweep).} 80 runs total against the Anthropic Messages API: 40 Claude Sonnet~4 (mean wall time $\sim 388$~seconds/run, ${\sim}4.31$~hours total) and 40 Claude Haiku~4.5 (mean wall time $\sim 269$ seconds/run, ${/sim}2.99$~hours total), dominated by API round-trips rather than local compute. Total API cost \$93.90 (\$69.68 Sonnet + \$24.22 Haiku); (Table~\ref{tab:appendix-cost}).

\textbf{Lean~4 verification.} \texttt{lake build} completes in ${\sim}10$~minutes on a standard laptop after the Mathlib cache is populated (approximately 3{,}496 compilation units, of which ours account for 5 files and ${\sim}300$ lines; the remainder is Mathlib).

\textbf{Preliminary / unreported compute.} Iterative prompt-engineering and per-edge enrichment-pipeline development during research consumed additional API calls and pentest pipeline runs that are not counted here; the $\$93.90$ figure above is strictly for the 80-run temperature sweep reported in Experiment~2. All three reported components (Experiment~1, Experiment~2, Lean verification) are runnable on a standard laptop once dependencies are installed.

\end{document}